# A Delay-tolerant, Potential field-based, Network Implementation of an Integrated Navigation System

Rachana Ashok Gupta, Ahmad A. Masoud, Mo-Yuen Chow,

*Abstract:* Network controllers (NC) are devices that are capable of converting dynamic, spatially-extended, and functionally-specialized modules into a taskable, goal-oriented group called: Networked Control System (NCS). This paper examines the practical aspects of designing and building a network controller that uses internet as a communication medium. It focuses on finding compatible controller components that can be integrated via a host structure in a manner that makes it possible to network, in real-time, a webcam, an unmanned ground vehicle (UGV), and a remote computer server along with the necessary operator software interface. The aim is to de-skill the UGV navigation process yet maintain a robust performance. The structure of the suggested controller, its components and the manner in which they are interfaced are described. Thorough experimental results along with performance assessment and comparisons to a previously implemented network controller are provided.

## nomenclature

iSpace: Intelligent Space.
QCFPT: Quadratic curve fitting path tracking
NCS: Networked Control System.
NC: Network Controller.
UGV: Unmanned Ground Vehicle.
HPF: Harmonic Potential Field.
GSM: Gain Scheduling Middleware.
LoG: Laplacian of Gaussian.
I(x, y): Raw top view image of the workspace.
V(x, y): Gradient Array from HPF.
R(x, y): Reference point corresponding to (x, y).
δL: Discrete look-ahead distance corresponding to one pixel.
$G_D$: Actual workspace distance in meters corresponding to one pixel in I(x, y).
v,ω: linear and angular velocity for the UGV.
m×n: Size of the workspace image I.
$x_a \times y_a$ : Actual workspace size in meters.
$C_{TM}$: Number of computations with template matching.
$C_{ED}$: Number of computations with edge detection and HPF.
k: Non-real time Computational performance improvement factor.
$\lambda_p$: Real time computational performance improvement factor.

## I. INTRODUCTION

Networked robotics offers a framework for coordination of complex mobile systems, thus providing the natural common ground for convergence of information processing, communication theory and control theory. It poses significant challenges, requiring the integration of communication theory, software engineering, distributed sensing and control into a common framework. Furthermore, robots are unique in that their motion and tasking control systems are co-located with the communications mechanisms. This allows the co-exploitation of individual subsystems to provide completely new kinds of networked or distributed autonomy. A physically dissociated navigation system, where the data processing, and controller modules could be freely-committed and physically located anywhere (network controller (NC))  has many advantages over a system where the navigation assets have to be physically related to a specific group of UGVs. An NC [1] [21-23] allows data to be efficiently shared. It is easy to fuse global information to take intelligent decisions over a large physical space and it  eliminates unnecessary wiring. It is also scalable because it is easy to add more sensors and UGVs with very little cost and without heavy structural changes to the whole system. Most importantly, an NC connects virtual space to physical space making task execution from a distance easily accessible (a form of tele-presence). With the advances in computer and communication technologies, these systems are progressively becoming easier to build.

One of the modes a network controller could operate in concerns the integration  of the assets a robotics agent needs for operation. In this case the bulk of the intelligence and processing needed for the robot to function is placed on a remote server that receives data of the environment of the robot and transmits guidance signal to the robot.  If properly designed, this layer may be used as an element for building another layer of networking. The higher layer is concerned with networking multi-robots so they will be able to share a common environment, even cooperate to collectively perform a certain task, e.g. RoboCup [46].

In order to actualize an NC that has certain capabilities, both the components and the networking scheme have to give rise to four basis functions which form the substrate of the capabilities an NCS is required to project (Figure 1). These basis functions are information acquisition, command, communication and control (C3I).

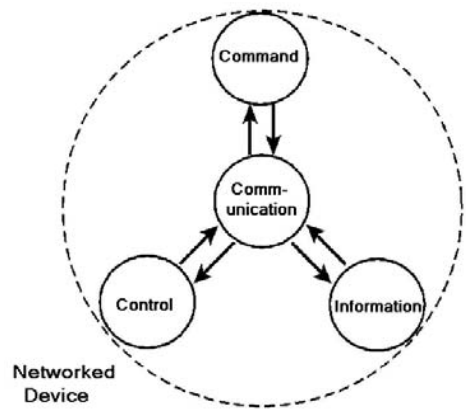

Figure 1: Basic functions in networked devices

A navigation controller that configures these  modules perform tasks independently yet together hence integrating them into one system. There is a wealth of techniques available for actualizing each one of the basis function modules. However, the norm seems to be: developing each module in isolation of the others by either placing, most probably restrictive, assumptions on the modules that can function with the one being developed, or assuming altogether, without providing a proof, that suitable modules do exist. An integrated view to the functioning of NCS that tackles design at the system level is still much in need.

A wide branch in literature [9] focuses on different control strategies and kinematics of the actuators/vehicles [20][32] Another research area concerning NCS studies the required network structure according to the network delays observed, packet losses and suitable protocols for control [33]. Techniques specifically focusing on navigation and path planning for autonomous robots using a predefined environment and obstacle models has also been an important subject of research for the past few decades. For example fast marching method by Sethian et al [26], potential field approach by Khatib [3-15], fuzzy logic based navigation system by Yen et al [34] are some of the popular techniques used for unmanned vehicles navigation [30][31]. Geometrical techniques like occupancy grids [35], topological mapping [36], voronoi diagrams using sonar, laser and visual sensors [15] are some of the techniques used to model the environment before path planning. While studying all these modules separately, it is highly unlikely to find a realistic command module that jointly takes into consideration the realization of an admissible control signal when converting a task and constraints on behavior into a group of reference guidance signals (a plan). There are few system architectures or middleware developed to put together a heterogeneous system [29][37]. These architectures focus mainly on modularization and abstraction of the system to achieve flexibility and scalability in design rather than on developing procedures and guidelines for selecting the controller components to enhance efficiency of each module separately as well as improving system performance as a whole. Same thing may be said about the information acquisition module. Many systems and algorithms such as template matching [25] and color recognition [39] have been developed using visual and other local sensing techniques to control ground and aerial vehicles autonomously. However, the suitability of the environment representation for use with the communication and command modules is rarely taken into consideration though it being the key point in any practical application of an NCS. The individualistic approach to design is highly likely to make an NCS overly expensive and complex. Also incompatibilities among the components could give rise to disruptive, hidden modes of behavior and seriously limit system performance. In [38] Brooks studied the requirements for an agent or a device to have reasonable chance of success operating in a realistic environment. He found that a device of such a sort must be intelligent to enable it to deal with novel situations it encounters the designer did not initially take into consideration; it must be situated as an integral part of the environment; it must be embodied and its behavior must be emergent.

In this work modules for realizing the basis functions needed for generating intelligent behavior are carefully selected and interlinked to realize a demonstrably-efficient, visually-servoed, networked controller for a differential drive UGV. The controller uses internet as a communication medium tying the data stream from an overhead camera to the processing and decision making unit that lies on a remote server back to the end-effector UGV. Although the controller focuses on networking resources needed by one UGV to operate, its design and the choice of modules yields a social controller that allows other UGVs using the same type of controller to share the same workspace (see section VI.F). The controller shows good robustness in presence of time delay. This among other things makes it possible to accommodate the channel coding process needed to guard against:
1- insertion of data in the communication channel
2- eavesdropping,
3- denial of message delivery by the recipient
4- data quality degradation caused by noise and other artifact.
Total cutoff of the guidance signal that is caused by the unavailability of the network can only be dealt with by the UGV reverting to the safe state of obstacle avoidance described in [2].

A major feature of the controller is the removal of the computationally-expensive, and error-prone image vison and interpretation module. The controller relies only on the raw output of a low-level vision, edge detection module to directly feed the planner. This provides the NC with the ability to handle image streams containing complex, unstructured components in a timely manner. Moreover, the planning and control modules are interfaced so that the NC exhibit high resistance to the unstructured and random delay internet induces in the visual servo loop. The modules are selected and networked in a manner that observes Brooks guidelines and minimizes inter-module incompatibility and conflict. Although there are no proofs that such guidelines can tackle the design of a general network controller, the authors found them useful in designing the network controller suggested in this paper. Experimental observations show that the networked modules synergetically interact to yield an NC with an intelligent, robust and emergent behavior that is satisfactorily situated in its environment.

Organization of the paper is as follows: Section II explains the NCS setup, design guidelines followed by the structural description of the test-bed iSpace which is used to implement the NC. Section III discusses the vision and planning modules used in building the NC along with the manner in which they are interfaced. In section IV the interface between the planning and the control modules is discussed. Section V compares the performance of the suggested NC to a previously implemented NC on the same test-bed. Thorough experimental results of the suggested NC are given in section VI and conclusions are placed in section VII.

## II. NC SETUP AND DESIGN GUIDELINES

This paper is the start of an effort to build a distributed resource NCS that provides a casual operator with means to effectively manage a group of mobile effector nodes that are operating in a static cluttered environment (Figure 2).

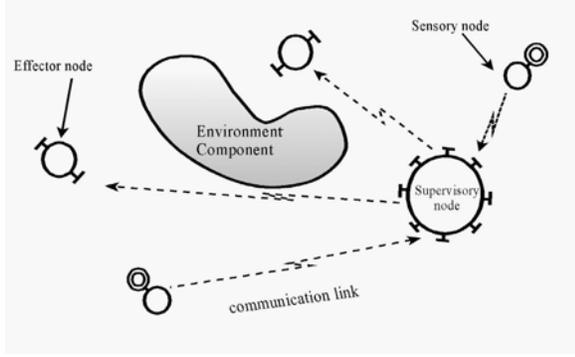

Figure 2: NCS basic layout.

In its fully developed form, it should be possible to deploy the effector nodes of the NCS to perform a variety of tasks that are set by a supervisory node. The components of the controller should be able to use different types of media to communicate among themselves or with the supervisory node. The components should also have adjustable autonomy with a behavior ranging from strict tele-operation by the supervisor based on an *a priori* available plan to a pure, on-the-fly, goal-driven self-organization. The system should respond to unfavorable circumstances such as loss of assets with graceful degradation in capabilities and performance instead of a total failure to carry-out the intended task.

An NCS prototype which we hope to be a starting point for building such a system is suggested in this paper. The proposed NCS is an intelligent, web-based, visual, servo system with a user/operator interface (Figure 3). It consists of a physical and a virtual environment. The physical environment is divided into two parts: the workspace environment which consists of a differential drive UGV operating in a stationary irregular clutter, a network camera recording the UGV and the workspace. The second part is the NC environment which consists of a communication channel and equipment, interface and data processing devices. The virtual environment consists of specialized software modules whose function is to construct a virtual representation of the environment (one is operator-centered and the other is planner-centered), command and decision making, reference generation and control.

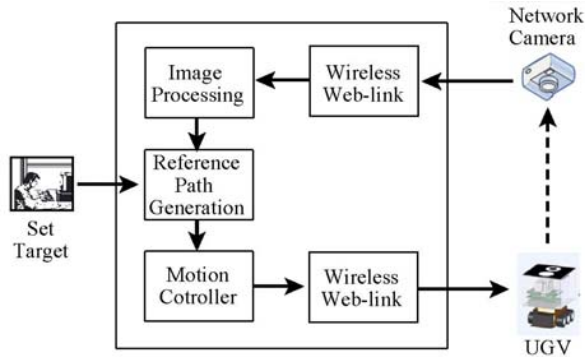

Figure 3: Prototype of the NBINS used as a test-bed -iSpace.

The NCS test-bed used here is called iSpace. iSpace is used to implement different algorithms and control strategies to compare and analyze the NCS performance. The iSpace platform [24] has the following components:

(i) An overhead network camera used as a global sensor to pass the visual information about the workspace to the main controller. Currently, this is the only source of sensory data the system is utilizing. Future work will consider using feed from local sensors onboard the UGV along with that from the webcam to enhance performance. The web camera collects the top-view image of the 2-D space of interest and sends it to the main controller for processing. The camera is mounted on the ceiling of the room (the indoor workspace) facing downwards capturing the 90o top view of the workspace as shown in the Figure 4 (details are in Table-1).

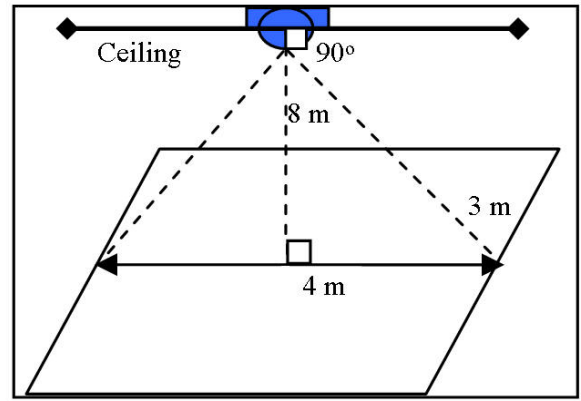

Figure 4: Single camera setup for the global workspace image

(ii) The NC with graphical user interface (GUI). The NC can be on any remote computer in the world with internet access. The main NC is connected to the global sensor and the actuator through a wireless channel. All modules and intelligence is thus located on the NC to process the top view image of the workspace from the network camera and generate the mobility control signal to be communicated to the UGV.

| Camera tilt angle | 0 |
|---|---|
| Workspace size per camera view ($x_a$ x $y_a$) | 3m x 4m |
| Size of the workspace image (m x n) | 320 x 240 |
| Camera height | 8 m |
| Camera image transfer rate (average) | 5 images /s |
| Image size (average) | 6KB |

Table 1: Camera implementation details and specifications

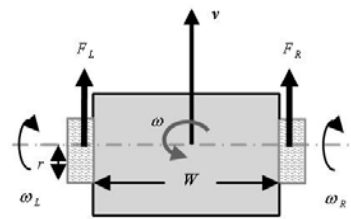

Figure 5: Model of a differential drive UGV

$$\begin{bmatrix} \omega_R \\ \omega_L \end{bmatrix} = \begin{bmatrix} 1/r & +W/2r \\ 1/r & -W/2r \end{bmatrix} \begin{bmatrix} \upsilon \\ \omega \end{bmatrix} \quad (1)$$

(iii) A differential drive UGV (Figure 5). The UGV does not have any local sensors on-board. It only receives linear speed (v) and angular speed (ω) from the NC on the wireless channel to control the motion. Eq.(1) is implemented in the UGV controller to drive the wheel motors.

(iv) The network used for communication is Internet. All the components are connected to the NC over 802.11b wireless channel. All the sensors and actuators at one location can be on LAN which is in turn connected to Internet for remote operations. The information and the control commands use UDP/IP for communication. UDP is less complex than TCP and saves the time required for the packet acknowledgment. This is an important required aspect for time sensitive systems since the control information carried by the dropped packet becomes obsolete to send again. As the delay on the actual network cannot be controlled or reproduced, a network delay generator (buffer implementation) is implemented for experimental purposes such that it can be inserted in the loop instead of the actual network to generate desired network delay in the system. Although it is not always the case that the dropped packets become obsolete to send again (re-transmissions, mainly with slow sampling systems, can be worth doing since they allow recovering the state of the system much faster than waiting for the next sample), there is no purpose in trying to retransmit a packet beyond a given packet deadline.

In this paper, we use the physical edges of the image stream from the camera as the source of information to represent the contents of the workspace. The HPF approach [3][4] to motion planning is used for generating the guidance signal for the UGV. The quadratic curve fitting path tracking (QCFPT) controller suggested in [20] is used for converting the guidance signal into a control signal the UGV may use for reaching a target point in the workspace.

iSpace is used as a test-bed to present the experimental results of this paper. It is also used to implement different algorithms and control strategies to compare and analyze their performance. We use edge detection to extract the contents of the workspace, HPF (Harmonic Potential Field) for guidance and a quadratic curve path tracking controller in [20] to generate the control signal. The GSM (Gain Scheduling Middleware) delay compensation module [22][23] is also used with the QCFPT. It is shown in the sequel that the above components can be networked together to yield a demonstrably-efficient NC. This particular choice of modules to realize the basis functions needed by an NC yields a compatible set of modules that satisfy the following conditions:
1) The format of the representation used for the workspace capture necessary and sufficient information about the environment for UGV navigation. It also has minimal size in order not to congest the communication channel.
2) Each module exhibits flexibility in terms of avoiding restrictive assumptions on the type of data it is required to process.
3) The format of the output from each module is compatible with that of the next module making it possible to practically accept the input in a raw form with almost a negligible amount of processing.
4) The modules, collectively, exhibit good resistance to artifacts the NCS may encounter such as delays in communication and processing as well as sensor noise.

We believe that these criteria are the key to building a system that yields satisfactory performance. The features possessed by the modules and the advantages in putting them together are discussed and demonstrated later in the paper.

### III. The Vision and Planning Modules

In this section a vision module utilizing a simple edge detector and the planning module which uses the harmonic potential field approach along with their interdependency are discussed.

#### A. Describing the contents of iSpace using edge detection:

Edge detection is a classic early vision tool used to extract discontinuities from the image as features. All the discontinuities which are more or less obstacle boundaries are represented by binary edges in the edge detected image of the UGV workspace. The detected edge image of the workspace is referred to as: the edge map. An edge map may be described using Eq. (2) as:

$$\begin{aligned} E(x_i, y_i) &= 1 \quad if \quad (x_i, y_i) \in \Gamma \\ &= 0 \quad if \quad (x_i, y_i) \notin \Gamma \quad \forall (i, j) \end{aligned} \quad (2)$$

where $E(x_i, y_i)$ is the image representing the edge map and $\Gamma$ is the set of boundary points for all obstacles in workspace. There are several edge detection techniques available in the literature. We decided to use the modified LoG (Laplacian of Gaussian) detector [18] [19]. It was proven in [18] that an LoG detector with a noise removal mechanism has many attractive properties. Though simple, it can be made robust when dealing with natural unstructured scenes having widely varying contrasts. The edge detector used here has two branches (Figure 6). One branch processes the workspace image I with a circularly symmetric approximated LoG operator to estimate the second derivative of the image. The zero crossing contours of this estimate are used to construct the potential edge map of the image.

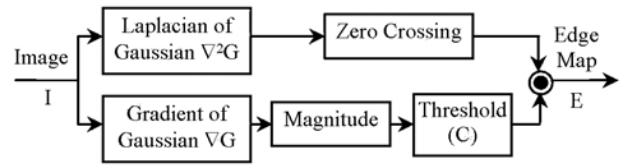

Figure 6: Complete edge detector block diagram

$$G(\sigma, x, y) = \frac{1}{2\pi\sigma^2} \exp(-\frac{(x^2 + y^2)}{2\sigma^2}) \quad (3)$$

$$\nabla^2 G(x, y) = \frac{\partial G}{\partial x^2} + \frac{\partial G}{\partial y^2} \quad (4)$$

$$I_{LoG} = I \otimes \nabla^2 G(x, y)$$

Such a map is extremely noisy and practically useless as shown in Figure 7 unless the contours corresponding to the physical underlying edges are appropriately filtered out [18].

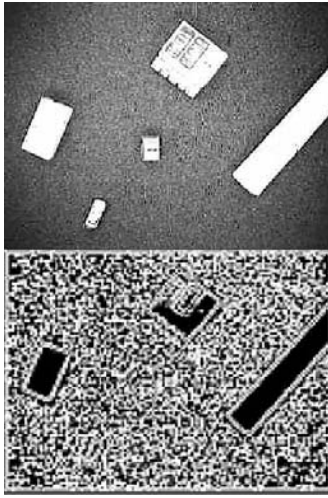

Figure 7: workspace image I and its zero crossing contour

The filtering adopted here is based on the strength of the edge, i.e. its contrast. If the contrast of an edge is above a certain threshold $\zeta$, the edge is accepted as authentic; otherwise, it is rejected as a noise induced artifact. Therefore, the second branch for the edge detection block is an approximated first derivative gradient operator Eq. (5) of 2-D Gaussian function Eq.(3). It is used to produce an estimate of the contrasts.

$$\nabla G(x, y) = \begin{bmatrix} \frac{\partial G}{\partial x^2} & \frac{\partial G}{\partial y^2} \end{bmatrix}^T \quad (5)$$
$$I_{GoG} = I \otimes \nabla G(x, y)$$

Figure 8 shows a few sample images from the actual iSpace setup with different backgrounds and obstacles along with their corresponding edge maps E(x, y).

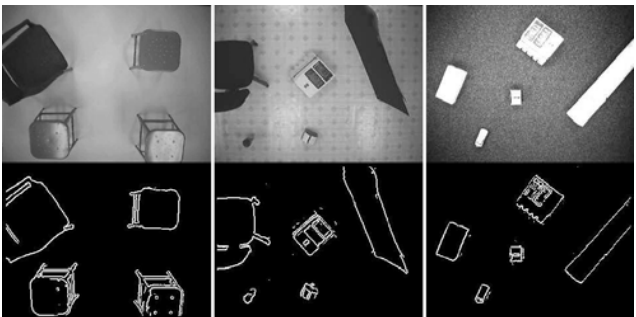

Figure 8: Output of the edge detector in figure-6.

B. 2-D HPF planner for path planning

The Harmonic function on a domain $\Omega \subset R^2$ for a continuously differentiable $\phi$ is a function which satisfies the following Laplace equation: Eq.(6) .

$$\nabla^2 \varphi = \frac{\partial \varphi}{\partial x^2} + \frac{\partial \varphi}{\partial y^2} = 0. \quad (6)$$

The use of potential field in motion planning was introduced by Khatib[2]. The use of harmonic functions for planning was suggested almost two decades ago [3][4] to generate a navigation path in robot workspace avoiding the spurious local minima in Khatib's settings. HPFs have proven themselves to be effective tools for inducing in an agent an intelligent, emergent, embodied, context-sensitive and goal-oriented behavior (i.e. a planning action). A planning action generated by an HPF-based planner can operate in an informationally-open (i.e. the planner does accept external information during the task execution phase) and organizationally-closed (i.e. the planner does not accept external assistance in making decisions) mode [10] enabling an agent to make decisions on-the-fly using on-line sensory data without relying on the help of an external agent. HPF-based planners can also operate in an informationally-closed (i.e. the planner does not accept external information during the task execution phase) , organizationally-open (i.e. the planner does accept external assistance in making decisions) mode which makes it possible to utilize existing data about the environment in generating the planning action as well as elicit the help of external agents . A hybrid of the two modes may also be constructed. In [11] vector-harmonic potential fields were used for planning with robots having second order dynamics. In [7] the HPF approach was modified to incorporate joint constraints on regional avoidance and direction. The decentralized, multi-agent, planning case was tackled using the HPF approach in [41]. A navigation HPF can be generated using VLSI chips both analog and digital [43-45].

A basic setting for generating the HPF from and edge map is:
$$\begin{aligned} & \nabla^2 \varphi(x, y) = 0 \quad x, y \in \Omega \\ \text{subject to:} \quad & \varphi(x, y) = 1 \quad (x, y) \in \Gamma \\ \text{and} \quad & \varphi(x, y) = 0 \quad (x, y) = (x_T, y_T). \end{aligned} \quad (7)$$

Where $\nabla^2$ is the Laplace operator, $\Omega$ is the workspace of the UGV ($\Omega \subset R^2$), $\Gamma$ is the boundary of the obstacles, and ($x_T$, $y_T$) is the target point. The obstacle free path to the target is generated by traversing the negative gradient of $\phi$ Eq.(8):

$$\begin{bmatrix} \dot{x} \\ \dot{y} \end{bmatrix} = -\nabla \varphi(x, y). \quad (8)$$

Convergence to the target point and avoidance of hazardous regions were proven in [7]

Figure 9 shows the result of the HPF planner after applying it to the edge maps of actual workspace images in iSpace. The arrows represent the negative gradient direction at each point converging to the destination point. Thus HPF converts the edge map into a region-to-point guidance function by accepting the edge data without any processing.

As mentioned in [3], a potential function in any connected component (within the closed boundaries) without a goal point will converge to a constant potential in that UGV configuration space hence leading to a vanishing guidance field. It is worth noting that all the connected components in the edge map represent the closed obstacle boundary. The successive relaxation method used in [3] to find the HPF is computationally efficient. The fixed number of iterations for a given workspace size and grid resolution, it makes the algorithm complexity to be O(k). As discussed in [40] HPF itself is a probabilistic measure of hitting the obstacles. As can be seen from the results obtained, successful integration of the LoG edge detector and the HPF planning module is achieved in a manner that uses minimal sufficient information needed to safely guide the robot to its target.

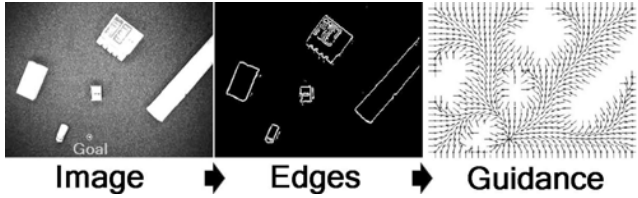

Figure 9: HPF-edge detector interface.

## IV The Planner-Controller Modules

In this section the interconnection between the HPF planning module and the path tracking module is described.

A QCFPT controller is implemented as the motion controller for the UGV to traverse the path generated by the planner from the source to the destination point. The basic principle of this control algorithm explained in [20] by Yoshizawa et al. is briefly described. A reference point is moved along a desired path so that the length between the reference point and the UGV is kept in some distance ($d_0$). Control (velocity) commands – linear speed ($v$, in cm/s) and angular speed ($\omega$, in rad/s) - are generated for the UGV to reach that reference position from the current position. The original algorithm was suggested as a solution to the path tracking problem. The path tracking controller is always trying to keep the UGV on the reference path by calculating the reference points along the path irrespective of the UGV's current position and distance from the reference path. The only input path tracking controller requires in each control loop iteration is the current coordinates of the UGV ($x_c, y_c, \theta_c$) and the reference point coordinates ($x_{ref}, y_{ref}$). In its current form this controller operates in a path-tracking mode. In the following, we describe an interconnection method with the HPF module that allows this controller to operate in the robust and efficient goal-seeking mode.

A. HPF planner and the motion controller integration

The QCFPT controller is used so far in case where the path to be tracked is already known. However, the gradient array of HPF has features that make it possible to use this method more efficiently without *a priori* knowing or calculating the path to be tracked. The 2-D gradient matrix from an HPF contains all the directional information required for the UGV to reach the goal from any point in the workspace (Figure 9). In other words, every point in the region is associated with a guidance vector that is directing it towards the next reference position leading towards the goal. This can be treated as the same reference position required by the quadratic curve controller.

Thus, the important Region to point guidance property of HPF makes it an efficient path planner to be combined with the quadratic controller without a reference path generation.

We need to compute the gradient array (V). $V_x$ and $V_y$ are the 2-D arrays of x- and y-components of $-\nabla\phi$ (negative gradient) respectively as described in Eq.(9) . The reference position ($x_R$, $y_R$) for each current position (x, y) is calculated from the gradient array of the HPF ($\nabla\phi$ ).Then each reference position $R(x_0, y_0) = (x_R, y_R)$ is calculated as shown in Eq.(10).

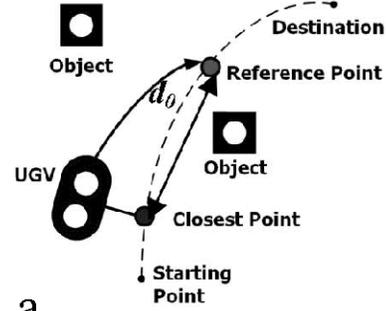

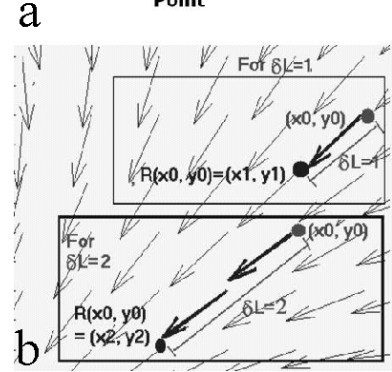

Figure 10(a): Path tracking control (b) HPF gradient directional array.

$$V_x = \frac{-\frac{\partial \phi}{\partial x}}{\sqrt{\left(\frac{\partial \phi}{\partial x}\right)^2 + \left(\frac{\partial \phi}{\partial y}\right)^2}} \quad , \quad V_y = \frac{-\frac{\partial \phi}{\partial y}}{\sqrt{\left(\frac{\partial \phi}{\partial x}\right)^2 + \left(\frac{\partial \phi}{\partial y}\right)^2}} \quad (9)$$

$$\begin{aligned} x_R &= x_0 + V_x(x_0, y_0) \\ y_R &= y_0 + V_y(x_0, y_0) \end{aligned} \quad (10)$$

We choose the reference position at a look-ahead distance ($d_0$) from the current position as shown in Figure 10(a). Let the discrete look-ahead distance be represented by $\delta L$ in pixels. If, $\delta L=2$, then for the current position ($x_0, y_0$), $R(x_0, y_0) = (x_2, y_2)$, as shown in Figure 10 (b). Thus for ($\delta L>1$), Eq.(10) is calculated in a loop of $\delta L$ iterations to get a reference point at a distance $\delta L$.

The accuracy of the path traced will be sensitive to $d_0$ as the quadratic curve controller will do curve fitting ($y = Ax^2$) between the current ($x_0, y_0$), and the reference point ($x_R, y_R$). The magnitude of the linear speed $v$ and the angular speed $\omega$ is proportional to $d_0$ between ($x_0, y_0$) and ($x_R, y_R$). Small $\delta L$ will keep the UGV trajectory close to the desired path. However the time to complete the path will increase because of the small $v$ and ω. With large δL (e.g. δL = 6), the quadratic curve controller will approximate the path between $(x_0, y_0)$ and $(x_6, y_6)$, which may not match the exact path generated by the HPF planner going through $\{(x_0, y_0), (x_1, y_1),...,(x_6, y_6)\}$. Thus, distance error increases and the probability of hitting nearby obstacles increases especially in the case of large network delays. (Figure 11, Figure 12, and Figure 13).

As described in [20], $d_0$ is calculated depending upon the curvature of the path (A) at that current position by Eq.(11a). Thus δL should be proportional to $d_0$, Eq.(11a). And then ν and ω are calculated using Eq.(11b). To choose the reference point to start with, we take δL as its minimum value i.e. δL = 1. Thus $(e_x, e_y, e_\theta)$ the error vector is calculated for $(x_c, y_c, \theta_c)$ and $(x_{ref}, y_{ref})$. To fit a curve,

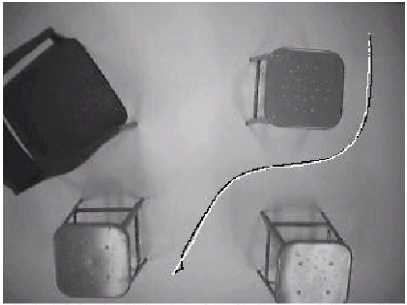

Figure 11: The effect of different values of look-ahead distances chosen. T is the total time to reach the destination.

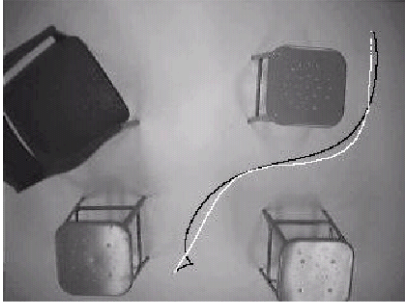

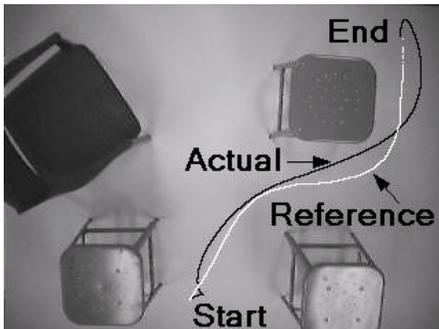

Figure 12: Large look-ahead distances with high network delay.

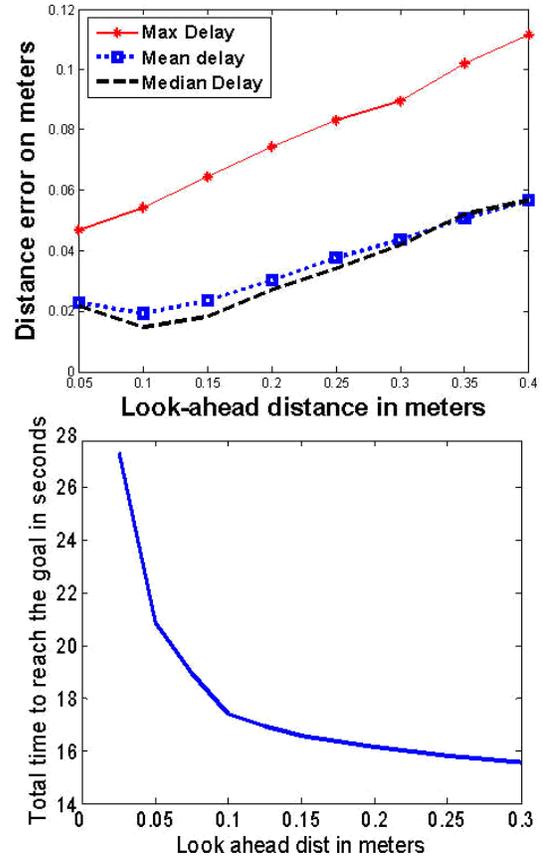

Figure 13: Distance error increases and total time required to reach the goal decreases with increase in δL.

$$y = Ax^2, \quad A = sign(e_x)\frac{e_y}{e_x^2}, \quad d_0 = \frac{d_{max}}{1+\beta\lfloor A \rfloor}, \quad \delta L = \left\lfloor \frac{d_0}{G_D} \right\rfloor \quad (11a)$$

$$K_n = sign(e_x)\frac{\alpha}{1+|A|}, \quad v = K_n, \quad \omega = 2A \cdot K_n. \quad (11b)$$

$G_D$ is a constant representing workspace resolution in meters/pixel. $G_D$ = 4/320 = 3/240 = 0.0125m with workspace image resolution as 320x240, the area covered by the network camera i.e. workspace size as (4m x 3m) and β is a positive constant the operator may use for tuning (here β=1). The above equations yield a δL that is inversely proportional to the curvature of the path. High curvature yields a small δL and a low UGV speed; while a straight path with small curvature yields a high δL and a high UGV speed. Thus dynamically choosing the look-ahead distance will deal a trade-off between the path tracking accuracy and the time required to reach the goal. Figure 14 shows the test results with dynamic look-ahead distance. Performance improvement in both the time to reach the destination and the tracking error with respect to the ideal path, is accomplished compared to the fixed look-ahead distance results in Figure 11. The flowchart describing the integration of the modules on the iSpace platform is shown in figure-15.

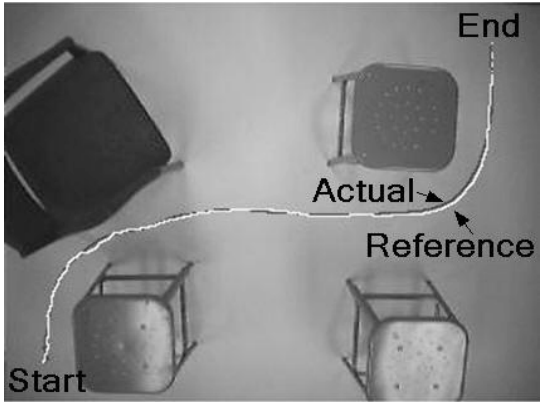

Figure 14: Test result with dynamic look-ahead distance.

## V COMPARISONS

The important modules in the iSpace are: obstacle registration, path planning, motion controller modules for the UGV and the network delay compensator [22][23]. While we implemented and presented results with the edge detection and HPF based structure for the integrated system, we also implemented the same NC with the popular template matching based obstacle recognition and fast marching based path planning modules integrated with the same quadratic curve fitting path tracking controller. The Following part of this section describes the implementation of template matching and fast marching method in iSpace and its comparison with the edge detection and HPF based structure. We compare the previous modules with the proposed ones and examine the performance improvement in iSpace as a whole.

### A. Template matching for content recognition

Template matching can be used to extract the information from the network camera image to know the location of the obstacles, position and orientation of the UGV in the workspace. In this case, we pasted circular templates on top of obstacles for recognition (circles are invariant to rotations). The camera is kept at a known constant height h over the workspace. The size of the templates on the top of the UGV and the obstacles are also kept constant allowing use of a template image with a fixed resolution (20×20) size. i.e. no zoom factor is considered as it increased the computations required by n-fold given that n is the number of zoomed images considered for matching. The NC does not have any information about the size of the obstacle. Therefore, we draw a circular safety margin of radius $r_{safe}$ around the obstacle. The area inside the safety margin is considered as part of the obstacle by the path planning module. Details about the template matching algorithm implemented in iSpace may be found in [24].

### B. Comparison of edge detection with template matching

Although the use of template matching is a standard practice in visual servo systems, there are many inherent factors in this approach that could limit performance. Some of these factors are:

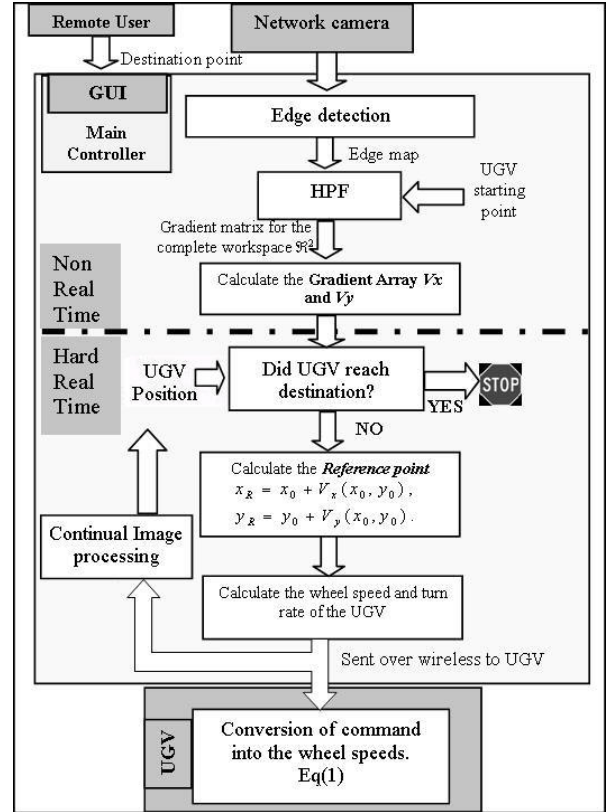

Figure 15: Complete flow diagram of the network based integrated system with edge detection and HPF planner.

(1) Restrictions on the contents of the environment: The obstacles recognized by template matching have to be a part of an *a priori* known set of obstacles with predefined size, shape, color and with predefined templates. These restrictions are removed by using edge detection.

(2) Definite knowledge of boundaries of obstacles: With template matching, the main controller only knows the location of the center of the template on the obstacle without having any quantitative and definite knowledge about the size and shape of the obstacle required for successful and guaranteed obstacle avoidance. The edge detector marks the boundaries distinctively. Thus the path planner directly receive from the edge detector better information about the contents of 2-D space of the UGV (Figure 16),

(3) Computational complexity: The major parts of the calculations involved in both template matching and edge detection algorithms are convolution with the template image and convolution of the edge kernel with the black and white workspace image. The size of the template image is 20 x 20, and edge kernel size is 7 x 7.

$C_{TM}$ = ((320- 20) x (240- 20)) x (20 x 20) = 26400000.
$C_{ED}$ = (320-7 x 240-7) x (7 x 7) = 3573521.
The factor k by which Edge detection is faster than the template matching can be estimated as shown in Eq.(12) :

$$k = \frac{C_{TM}}{C_{ED}} = \frac{264000000}{3573521} \approx 7.3 \qquad (12)$$

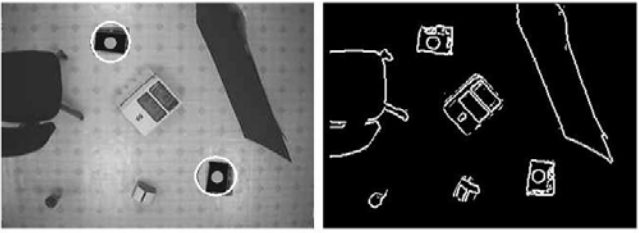

Figure 16: Template matching and Edge detection outputs for the same workspace image.

C. Path planning using fast marching method

The input to the path generation modules is the position and orientation of the UGV ($x_u$, $y_u$, $\theta_u$) and obstacles' positions received from of the template matching module, and the UGV's destination ($x_d$, $y_d$) from the GUI set by the user. The output of this module is an obstacle-free path composed of a series of points from the initial UGV position to the destination. A Dijkstra-like method named the fast marching method was applied to solve the optimal path generation problem [26].

D. The Fast marching path planner vs. the HPF path planner

(1) Problem description: The reference path calculated by the fast marching method is the function of the source point, destination point and the obstacle map. The point of consideration is that being a NCS (e.g. iSpace), the UGV might wander off from the path in due course of movement due to network delay. In such case, it might prove practical and efficient for the UGV to go to the next reference position which might not lie on the previously calculated reference path. Thus once the UGV is not on the path there is no guarantee that the quadratic curve fitted between the UGV (off the path) and the reference position (on the path) considers the obstacle avoidance. This creates a requirement for each point in space to have a reference point toward the destination which will have built-in knowledge of the obstacle boundary presence instead of a reference path. With HPF planner, the workspace surface image is converted into a region-to-point guidance function that is dependant on the destination point and the (obstacle map) edge map. Thus each point in the region has a reference point associated with it towards the destination and avoiding the obstacle making sure that quadratic curve fitted between the UGV (anywhere in the region) and its reference point is avoiding the nearby obstacle.

Although the reference point is chosen on the path (Figure 17(a)), it increases the probability of hitting the corner of the obstacle. The reason is that the proximity of the obstacle was not being considered. In Figure 17(b), the reference point is chosen at the same distance from the current position, but the current position being near the obstacle the Dirichlet's setting generates the gradient value driving the UGV away from obstacle. Since the reference point is calculated from the gradient point, it is naturally pushed away from the obstacle.

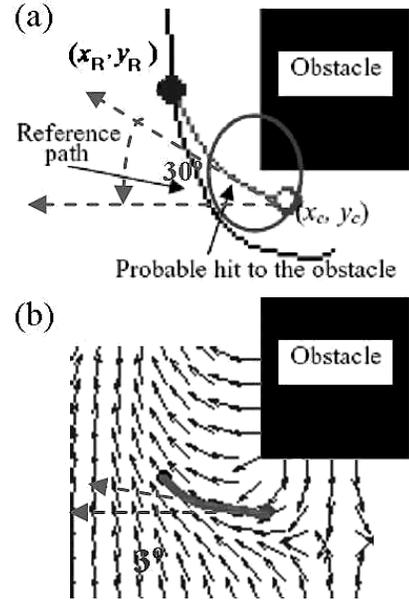

Figure 17: Behavioral comparison between "path tracking" and "goal seeking" problem.

(2) Complexity of the fast marching is $O(N \cdot \log N)$. On the other hand, HPF consists of simple averaging operation on the edge array a fixed number of times (say M) therefore making it an $O(M)$ algorithm. This comparison clearly shows that HPF planner is computationally more efficient than the fast marching method as N increases.

(3) The following section explains the performance improvement in real-time computation time with the new structure of HPF planner and motion controller (goal seeking approach) with respect to the fast marching method and motion controller (path tracking approach). In the fast marching approach, the complete path coordinates are pre-calculated before the UGV movement starts. The pre-determined reference path has Np discrete points, $\{(x_0, y_0), (x_1, y_1), ..., (x_{Np}, y_{Np})\}$, from source to the goal. At each sampling instant $t_i$, the current position of the UGV ($x_0$, $y_0$) is determined using template matching. Each of the Np points on the pre-determined path is checked to find out which point is closest to ($x_0$, $y_0$). The reference point ($x_R$, $y_R$) on the path is determined such that it is at a set distance ($d_0$) from the closest point. Thus, at each $t_i$, it will take Np number of real time checks on the path to find out the closest point and the corresponding reference position. When HPF is combined with the quadratic controller, at each $t_i$, the reference point computation in real time will take $\delta L$ additions. This decreases the real-time computation cost and the time by a factor of: $\lambda_p = N_p/\delta L$, Np being the total number of points from source to destination on the path and $\delta L$ being the discrete look-ahead distance towards the destination, Eq.(13) will always be satisfied.

$$N_p \geq \delta L \quad \therefore \quad \lambda_p \geq 1 \quad \quad (13)$$

The longer the path (Np) is, the more efficient the HPF planner will be.

## VI. Experimental testing of the suggested NCS.

The capabilities and features of the suggested system are demonstrated using five case studies. In case of the HPF planner, the reference path from source to destination is calculated and compared with the actual path of the UGV. The distance between the UGV's position and the ideal path is calculated at each time $t_i$ with respect to the time delay. This difference is called the distance error. It shows how close the actual path is to the ideal path. The total time to reach the destination is also recorded. The distance error and the total time are the main criteria used for performance evaluation.

### A. Case-1: Different obstacle and background cases

Figure 18 shows satisfactory performance of the suggested method with different number, shapes and sizes of obstacles in the workspace. We observe that all the turns around the obstacle are at enough safe distance. Without tracking the ideal path, the reference array approach yields the average and the maximum distance error of 2 to 5 cm respectively, which is 1% error with respect to the dimension of the space.

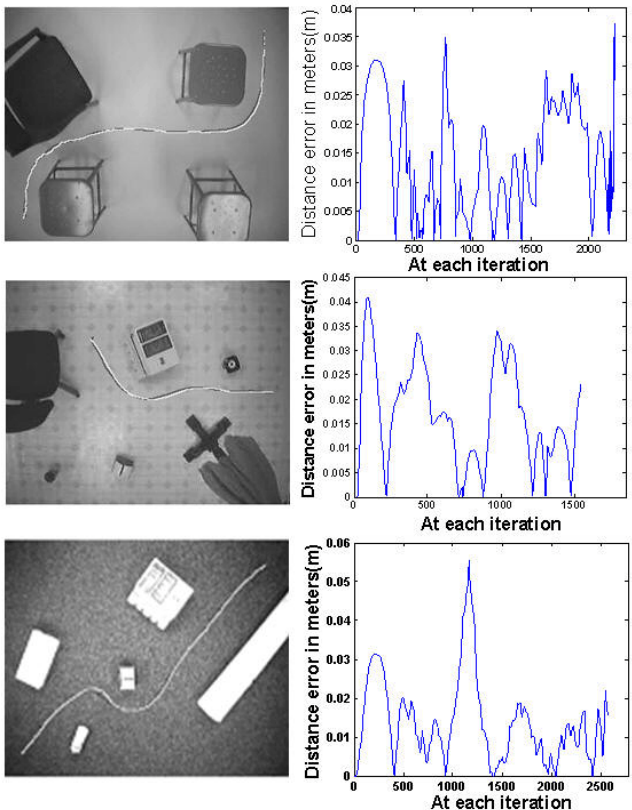

Figure 18: Workspace images and the graph of distance error vs time

### B. Case-2: Network and Processing delays

Generally speaking, the network delay is in the order of few hundreds of millisecond on the Internet. Typical delays observed in US continent are 0.1 s to 0.6 s. We however tested the functionality for a network delay range of 0.1 s to 1.2 s. Figure 19 shows that the distance error increases exponentially with increase in network delay. For the average network delay of 0.3 seconds, we have a mean distance error of 1cm and a maximum error less than 5 cm. Even with delays as large as 1.2 sec, the maximum error is 20 cm (4% of the diagonal length of workspace area).

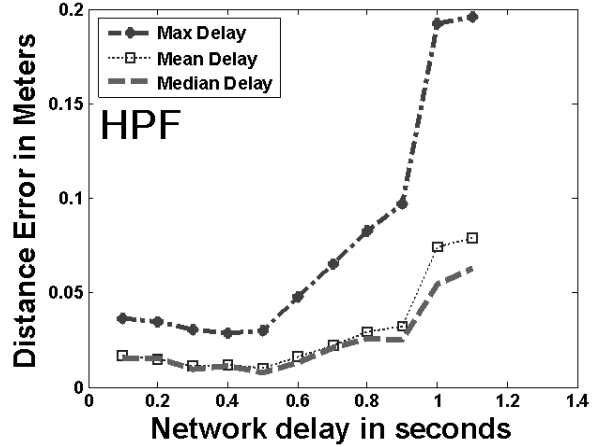

Figure 19: distance error measures vs Network delay.

### C: Case-3: Path quality

Figure 20 and Figure 21 demonstrate that the path from the fast marching-based NC exhibit more variations in the curvature graph (sudden turns) unlike HPF which provides a smoother path. In other words, the path obtained from the suggested NC (HPF-based approach) is more dynamically friendly to traverse than the path obtained from the FM-based approach. Figures 22 and 23 also clearly show that the path generated by the suggested approach is less susceptible to delay-induced errors than the one from the FM-based approach. It can be clearly seen that a goal-seeking mode significantly improves performance.

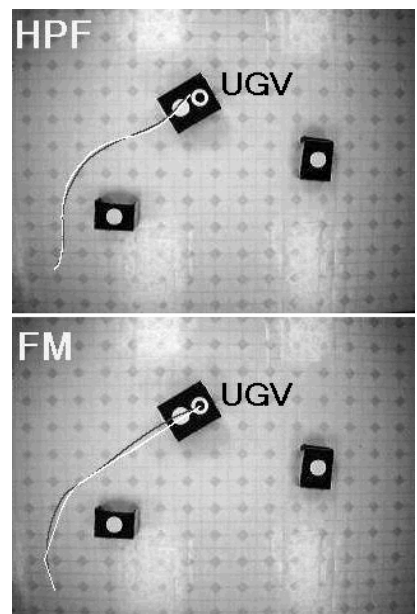

Figure 20: HPF planner versus fast marching (FM) planner.

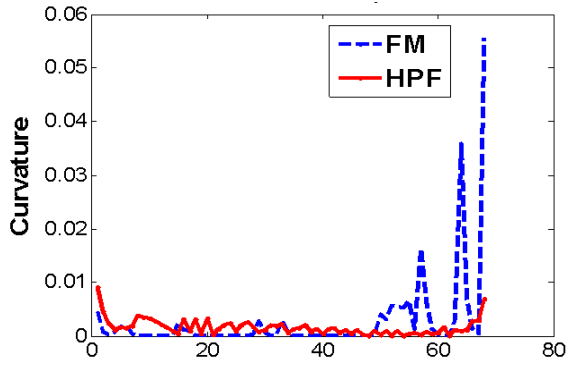

Figure 21: Curvatures for Figure 20

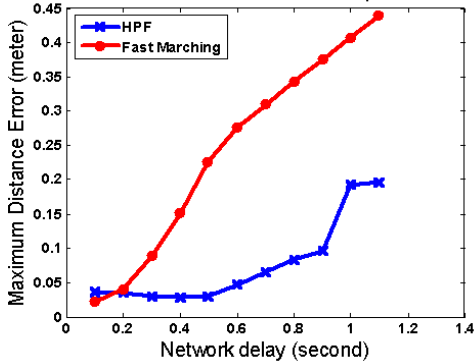

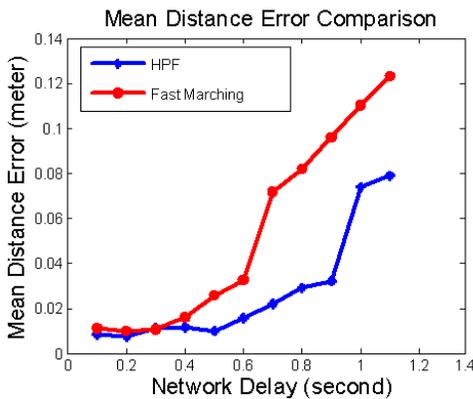

Figure 22: Maximum and mean distance error vs. network delay comparing fast marching and HPF planner.

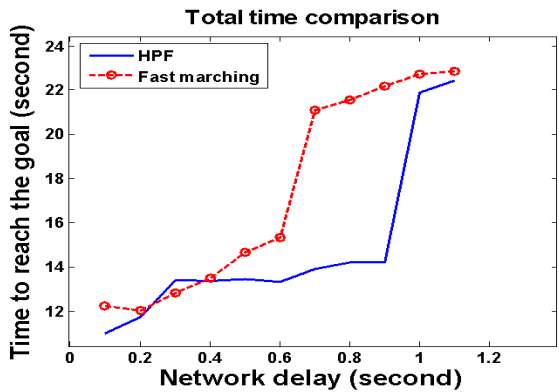

Figure 23: Total time vs. network delay comparing fast marching and HPF.

D. Case-4: Special cases:

An interesting, intelligent behavior is observed during the experiments when the UGV is initially oriented in an opposite direction to the gradient guidance field. The UGV simultaneously combines the basic behaviors: driving in reverse, turning around, staying away from the obstacle, and proceeding towards the target to synthesize a human driver-like maneuver that treats the space as a scarce resource.

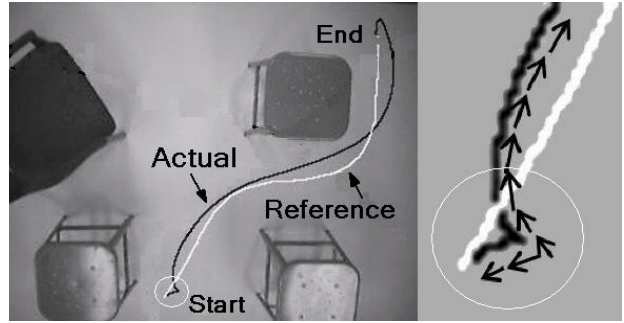

Figure 24: Direction reversal maneuver. Arrows show the UGV orientation.

HPF displays an intelligent behavior when the goal point chosen is unreachable due to a barrier in the workspace. We deliberately put an obstacle so that the workspace is divided into two separate regions as shown in Figure 25. We kept the UGV in one region and chose the destination point in another region that is unreachable by the UGV due the wall like obstacle dividing the workspace into isolated regions. The complete region around the UGV is pulled down at an equal high potential. This created a zero gradient (flat region) around the UGV. Vx and Vy are both zero giving the next reference point for the UGV to be the same point as per Eq.(10) Therefore the UGV does not move from its position as if knowing beforehand that the goal point is unreachable. (Figure 25)

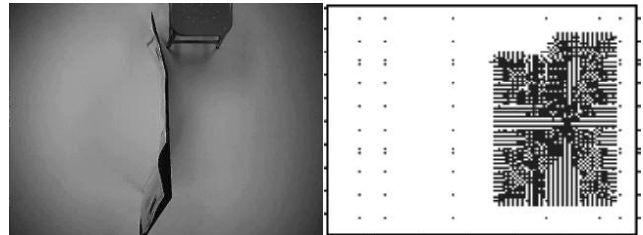

Figure 25: Image with a barrier separating two regions..

E. Case-5: Robustness to lighting conditions:

Reliable edge detection is the backbone of the vision module. The harmonic potential planner can use a raw edge map to generate the guidance field without any need to interpret the map's contents (i.e. the contents of the map are unknown to the operator). There are several difficulties that could arise when the edges of a non-synthetic image are to be detected. Noise, shadows, variable texture or contrasts in the image could all cause false edge components. We performed a few experiments for different lighting and contrast conditions for one image. In Figure 26, we observe that we loose edges as well as get extra edges as lighting condition changes. However HPF planner successfully generates paths from source to destination in all the cases (black line showing the generated path). The fact that the NC did not show much sensitivity to missing edges is not luck. It has to do with the choice of the planning module. Harmonic potential is intimately connected to collision probability [40]. The HPF planner guides motion to reduce this probability. So if part of the edge contour is missing, the probability of collision at the missing part (i.e. the harmonic potential) is still high and the planner will attempt to guide motion away from it. These are just preliminary results to demonstrate that the HPF planner has a considerable level of tolerance to artifacts in the edge map or to loss of authentic edges and the appearance of false ones.

approach has. The HPF planer is naturally decentralized making it a valid guidance protocol in a multi-agent environment where each agent uses the HPF approach for guidance. This means that NCs similar to the one suggested in this paper can act independently on robots sharing the same workspace yet collectively enable each UGV to safely proceed to its target, Figure 27.

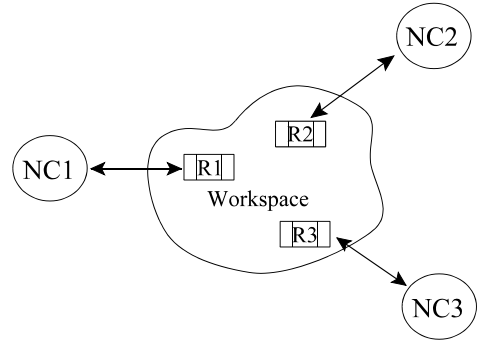

Figure-27: The suggested NC is naturally decentralized.

Despite the fact that the controllers are not aware of each others' presence, the overall control action regulating the group of robots in the workspace is a distributed, evolutionary, self-organizing one.

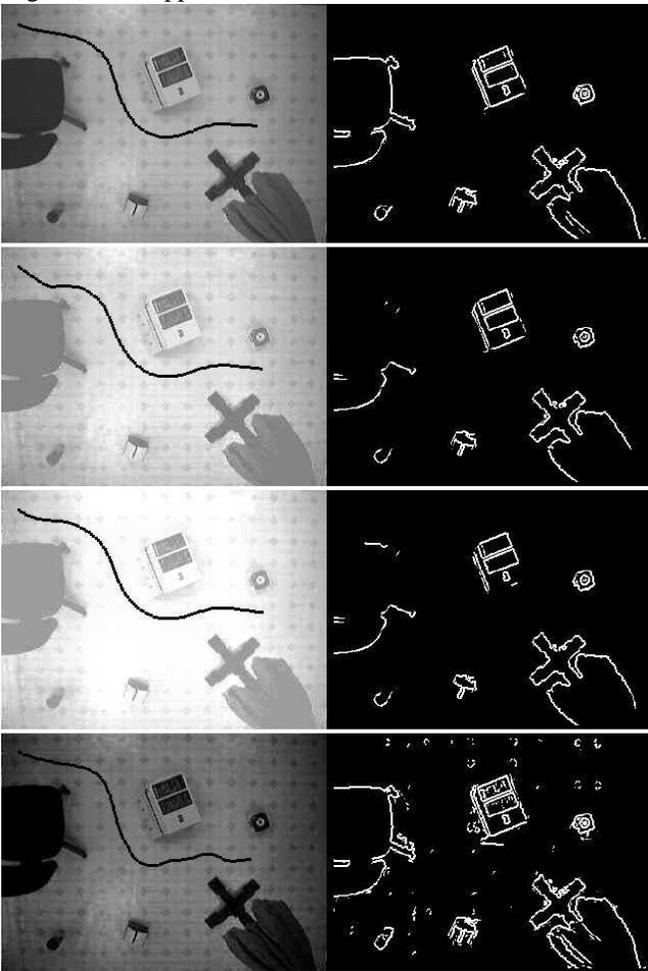

Figure 26: Experiments with different lighting conditions

### F. A note on the multi-UGV case:
While the focus of this paper is on networking the modules for a single UGV, the suggested controller has the ability to operate in a multi-UGV environment. This ability stems from fundamental capabilities which the harmonic potential field

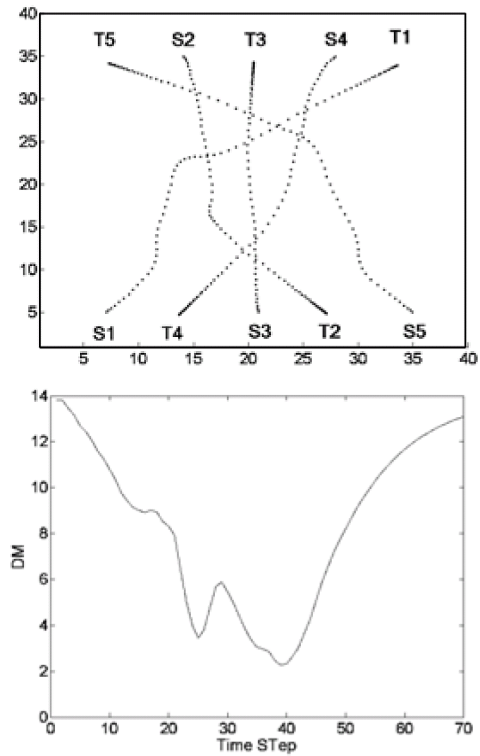

Figure-28: HPF planners have full awareness of the other agents in the environment.

Figure-28 shows five robots controlled individually by a harmonic potential field planner. Each robot is required to move from a start point (S) to a target point (T). The minimum distance (DM) between the robots is plotted for all time. As can be seen, the decentralized collective of harmonic planer successfully generates a safe trajectory for each robot to its target. In Figure-29 a close to a catastrophic failure situation is created. The individual planners neglect the presence of all the robots in the workspace except the one closest to the robot they are steering. The simulation results are shown below. The group kept functioning normally as if nothing happened.

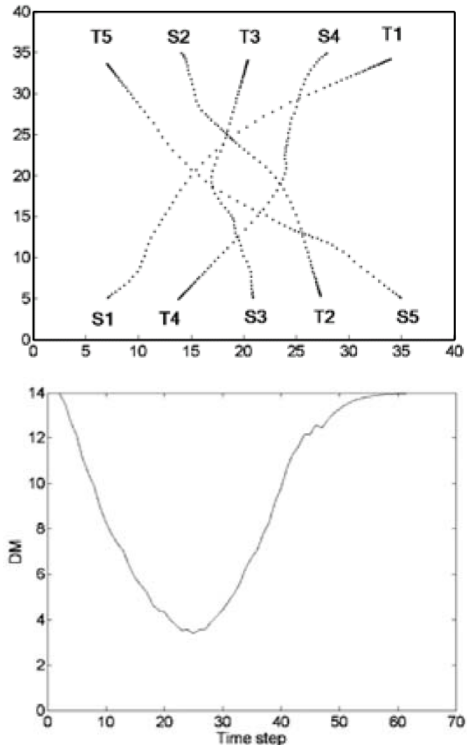

Figure-29: Each HPF planner is only aware of its nearest neighbor.

## VII. CONCLUSION, DISCUSSION AND FUTURE WORK

This paper has successfully shown through experimental results that edge detection, an HPF planner and a network based quadratic controller can be networked to create an efficient and delay-tolerant integrated network controller. The edge detection and HPF-based planning modules satisfy several important features that are central to the integration of the system. This gives more generality and flexibility to the UGV environment which is not the case with template matching. The efficiency of the suggested approach is evident from the comparison with the fast marching method. Moreover, the memory storage needed by the method is negligible (order of Kilobytes) and the computations can be carried-out in-place.

We observe that the ratio of the number of edge pixels (information pixels) to the total number of pixels in the image is always less than one. In case of iSpace we observed it to be on the average 5% of the whole image pixels. Current results for the NC are obtained for whole frame transmission. If an IP camera or hardware with local processing is used to transmit only the edge pixels, a much superior performance compared to the one we already have is expected. The NC will use lesser bandwidth than required for the whole frame transmission. As for scalability:

1- All the components of the NCS can be built/implemented using cheap, easily available off-the-shelf hardware components. Thus adding more cameras, UGVs or NCs with different network addresses to the workspace is a very quick and easy job ensuring the scalability.

2- It is possible to add any number of cameras in the ceiling to increase the workspace area coverage. As the tilt angle is zero and the cameras are centered in the workspace, they can be spaced in the ceiling in such a way that they form an array of cameras to cover a larger workspace. The edge maps from the different cameras can be easily combined using a logic OR operation.

3- Currently the NC is programmed to handle only one agent (UGV) however, HPF can be naturally extended to control multi-robot with very little modification to the existing structure as explained in [41]. Each agent can be assigned one NC, all other agents can be treated as moving obstacles.

The physical evidence regarding the modules selected for building the NC and the manner in which they are networked makes us believe that the suggested structure can be extended with reasonable effort to more challenging applications. There is still a lot of work that needs to be done in order to extend the capabilities of the suggested approach. For example, the robot end-effector is expected to carry a payload. While considering the kinematic aspects only at the high-level control stage is sufficient for many practical cases, accommodating dynamics at the high-level control stage will have to be addressed. Also, the incorporation of coding in the communication channel to protect data fidelity has to be incorporated in a manner that does not degrade the performance of the system